\documentclass[letterpaper]{article} 
\usepackage[draft]{aaai25}  
\usepackage{times}  
\usepackage{helvet}  
\usepackage{courier}  
\usepackage[hyphens]{url}  
\usepackage{graphicx} 
\usepackage{natbib}  
\usepackage{caption} 
\usepackage{algorithm}
\usepackage{algorithmic}

\usepackage{newfloat}
\usepackage{listings}
\DeclareCaptionStyle{ruled}{labelfont=normalfont,labelsep=colon,strut=off} 
\floatstyle{ruled}
\newfloat{listing}{tb}{lst}{}
\floatname{listing}{Listing}
\usepackage{multirow}
\usepackage{booktabs}
\usepackage{array}
\usepackage{xcolor}
\usepackage{amssymb}
\usepackage{amsmath}

\definecolor{codegreen}{rgb}{0,0.6,0}
\definecolor{codegray}{rgb}{0.5,0.5,0.5}
\definecolor{codepurple}{rgb}{0.58,0,0.82}
\definecolor{backcolour}{rgb}{0.95,0.95,0.92}
\lstdefinestyle{pnpxai}{
    backgroundcolor=\color{backcolour},
    commentstyle=\color{codegreen},
    keywordstyle=\color{magenta},
    numberstyle=\tiny\color{codegray},
    stringstyle=\color{codepurple},
    basicstyle=\ttfamily\tiny, 
    breakatwhitespace=false,         
    breaklines=true,                 
    captionpos=b,                    
    keepspaces=true,                 
    numbers=left,                    
    numbersep=5pt,                  
    showspaces=false,                
    showstringspaces=false,
    showtabs=false,                  
    tabsize=2
}
\title{PnPXAI: A Universal XAI Framework Providing Automatic Explanations\\Across Diverse Modalities and Models}
\author {
    Seongun Kim\textsuperscript{\rm}\thanks{Equal contribution},
    Sol A Kim\textsuperscript{\rm}\footnotemark[1],
    Geonhyeong Kim\textsuperscript{\rm}\footnotemark[1],
    Enver Menadjiev\textsuperscript{\rm}, \\
    Chanwoo Lee\textsuperscript{\rm},
    Seongwook Chung\textsuperscript{\rm},
    Nari Kim\textsuperscript{\rm}\footnotemark[2],
    Jaesik Choi\textsuperscript{\rm}\thanks{Co-corresponding}
}
\affiliations {
    \textsuperscript{\rm}Kim Jaechul Graduate School of AI, KAIST\\
    \{seongun, ksola, geonhyeong.kim, enver, shiningstone23, seongwook.chung, nari.kim, jaesik.choi\}@kaist.ac.kr
}

\usepackage{bibentry}

\begin{document}

\maketitle

\begin{abstract}

Recently, post hoc explanation methods have emerged to enhance model transparency by attributing model outputs to input features. However, these methods face challenges due to their specificity to certain neural network architectures and data modalities. Existing explainable artificial intelligence (XAI) frameworks have attempted to address these challenges but suffer from several limitations. These include limited flexibility to diverse model architectures and data modalities due to hard-coded implementations, a restricted number of supported XAI methods because of the requirements for layer-specific operations of attribution methods, and sub-optimal recommendations of explanations due to the lack of evaluation and optimization phases. Consequently, these limitations impede the adoption of XAI technology in real-world applications, making it difficult for practitioners to select the optimal explanation method for their domain. To address these limitations, we introduce \textbf{PnPXAI}, a universal XAI framework that supports diverse data modalities and neural network models in a Plug-and-Play (PnP) manner. PnPXAI automatically detects model architectures, recommends applicable explanation methods, and optimizes hyperparameters for optimal explanations. We validate the framework's effectiveness through user surveys and showcase its versatility across various domains, including medicine and finance.
\end{abstract}

\begin{links}
    \link{Code}{https://github.com/OpenXAIProject/pnpxai}
    \link{API Doc.}{https://openxaiproject.github.io/pnpxai/}
    \link{Demo}{https://openxaiproject.github.io/pnpxai/demo}
\end{links}

\section{Introduction}

In recent years, various post hoc explanation methods have emerged as a promising approach for enhancing the model transparency. These methods aim to provide insights into the decision-making process of complex neural networks by attributing the model's output to its input features. Techniques such as gradient-based methods \cite{srinivas2019full, smilkov2017smoothgrad, adebayo2018sanity}, relevance propagation methods \cite{bach2015pixel, nam2020relative}, and model-agnostic methods \cite{ribeiro2016should, lundberg2017unified} have gained significant attention for their ability to generate interpretable explanations. These methods are crucial for building trust in AI systems, especially in mission-critical applications like healthcare \cite{bassi2024improving}, finance \cite{misheva2021explainable}, and robot manipulation \cite{kim2021explaining}.

However, applying these input attribution methods is challenging due to several factors. Some methods are tailored to specific neural network architectures, like assuming a series of convolutional layers followed by linear layers, which limits their generalizability across different models \cite{selvaraju2017grad}. Others require operations specific to certain layers, adding complexity to their application on complex architectures \cite{bach2015pixel, nam2020relative}. Additionally, some methods assume extra operations based on data modalities \cite{sundararajan2017axiomatic}.

\begin{figure*}[!ht]
    \centering
    \includegraphics[width=\linewidth]{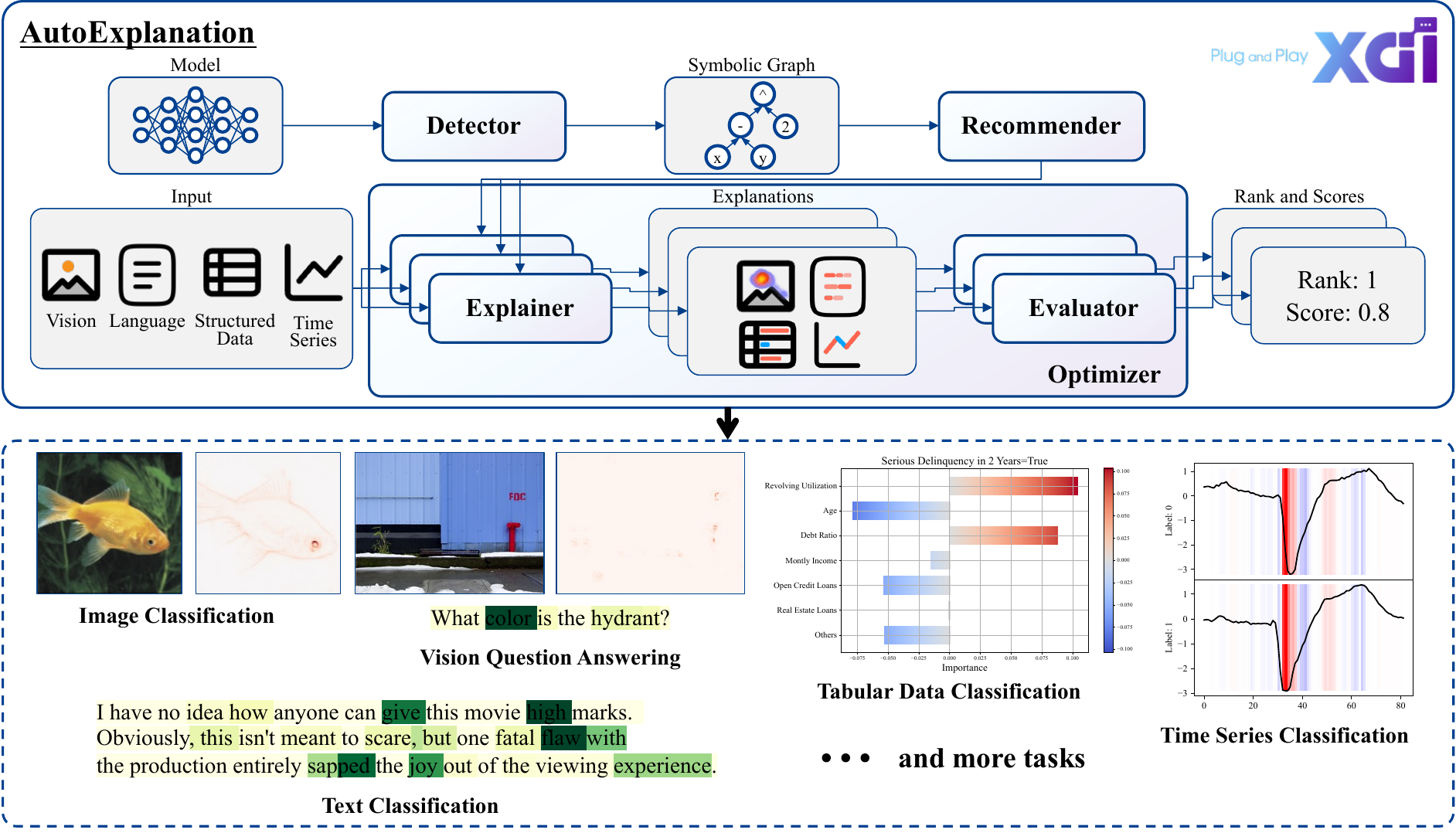}
    \caption{\textbf{Overview of the proposed framework, PnPXAI.} The detector module automatically identifies the provided neural network model architecture, which the recommender module uses to filter applicable explanation methods. The evaluator module then optimizes the explanation results through hyperparameter optimization before presenting them to end users.}
    \label{fig:pnpxai_frameweork}
\end{figure*}

\begin{table*}[!t]
    \centering\resizebox{.7\linewidth}{!}{
    \begin{tabular}{l||cccc|cc|ccccc}
    \toprule
        \multicolumn{1}{l||}{\textbf{Framework}} & \multicolumn{4}{c|}{\textbf{Modalities}} & \multicolumn{2}{c|}{\textbf{Methods}} & \multicolumn{5}{c}{\textbf{Modules}} \\
        & SD & V  & L  & TS & MA & MS & Det. & Recom. & Eval. & Opt. & AutoExp. \\ 
    \midrule
    \textbf{PnPXAI (ours)}  & \checkmark & \checkmark & \checkmark & \checkmark & \checkmark & \checkmark & \checkmark & \checkmark     & \checkmark    & \checkmark   & \checkmark       \\
    Captum$^1$  & \checkmark & \checkmark & \checkmark & \checkmark & \checkmark & \checkmark &    &        & \checkmark    &      &          \\
    OmniXAI$^2$ & \checkmark & \checkmark & \checkmark & \checkmark & \checkmark & \checkmark &    &        &       &      &          \\
    AutoXAI$^3$ & \checkmark &    &    &    & \checkmark &    &    & \checkmark     & \checkmark    & \checkmark   & \checkmark       \\
    Xaitk-Saliency$^4$   &    & \checkmark &    &    & \checkmark &    &    &        &       &      &          \\
    OpenXAI$^5$ & \checkmark &    &    &    & \checkmark & $\triangle$  &    &        & \checkmark    &      &          \\ \bottomrule
    \end{tabular}}
    \caption{\textbf{Comparision between PnPXAI and other XAI frameworks.} MS for OpenXAI is marked as $\triangle$ because it doesn't support explainers that require layer-wise operations. $ ^1$Captum~\cite{kokhlikyan2020captum}. $ ^2$OmniXAI~\cite{yang2022omnixai}. $ ^3$AutoXAI~\cite{cugny2022autoxai}. $ ^4$Xaitk-Saliency~\cite{hu2023xaitk}. $ ^5$OpenXAI~\cite{agarwal2022openxai}. Abbreviations: SD - Structured Data, V - Vision, L - Language, TS - Time Series, MA - Model-Agnostic, MS - Model-Specific, Det. - Detector, Recom. - Recommender, Eval. - Evaluator, Opt. - Optimizer, and AutoExp. - Automatic Explanation.}
    \label{tab:intro_comparision}
\end{table*}


Recent efforts have attempted to address the limitations by providing a set of implementations of explanation methods with additional features~\cite{kokhlikyan2020captum, agarwal2022openxai, cugny2022autoxai}, collectively known as explainable artificial intelligence (XAI) frameworks.
However, these frameworks have their own constraints (see Table~\ref{tab:intro_comparision}). They face challenges when applied by end users, including (1) limited flexibility for diverse and custom models due to hard-coded implementations~\cite{agarwal2022openxai, hu2023xaitk}, (2) a restricted number of supported XAI methods because of layer-specific requirements~\cite{agarwal2022openxai, hu2023xaitk, cugny2022autoxai}, and (3) sub-optimal explanations due to the lack of integrated evaluation and optimization phases~\cite{kokhlikyan2020captum, yang2022omnixai}. Finally, yet equally significant, most XAI frameworks lack user-friendly tools to help users select and optimize suitable algorithms for their own tasks, hindering the effective adoption of XAI in real-world applications.

To address these challenges, we introduce PnPXAI, a universal XAI framework that embraces diverse data modalities and neural network models for ease of use. PnPXAI is equipped with the capability to automatically detect the model architecture and recommend applicable explanation methods without necessitating an in-depth understanding from the end user. We validate the effectiveness of our framework through a user survey, demonstrating the satisfaction and usefulness of the PnPXAI framework. Furthermore, we showcase the versatility of our framework by presenting use cases across various domains and data modalities, including but not limited to medical image classification and fraud account detection.

\section{PnPXAI Framework}
\label{sec:pnpxai_framework}
\begin{table*}[!ht]
    \centering\resizebox{1.\linewidth}{!}{
    \begin{tabular}{l||>{\centering\arraybackslash}p{4cm}|>{\centering\arraybackslash}p{9cm}}
    \toprule
        \textbf{Method} & \textbf{Data Modalities} & \textbf{Architectures} \\ 
    \midrule

        LIME, KernelSHAP & V, L, SD, TS & Linear, Convolution, Recurrent, Transformer, Decision Trees \\ 
        Gradient, Gradient $\times$ Input  & V, L, TS & Linear, Convolution, Recurrent, Transformer \\ 
        Grad-CAM, Guided Grad-CAM  & V, TS & Convolution \\ 
        FullGrad, SmoothGrad, VarGrad & V, L, TS & Linear, Convolution, Recurrent, Transformer \\ 
        Integrated Gradients, LRP, RAP & V, L, TS & Linear, Convolution, Recurrent, Transformer \\ 
        AttentionRollout, TransformerAttribution & V, L & Transformer \\
    \bottomrule
    \end{tabular}}
    \caption{\textbf{Mapping table for the recommender module.} The recommender module filters the applicable explanation methods by intersecting the identified data modalities and model architectures by the detector module. The abbreviations in the data modalities column are as follows: V for Vision, L for Language, SD for Structured Data, and TS for Time Series.}
    \label{tab:mapping_table}
\end{table*}

To automatically detect the model architecture and recommend applicable explanation methods, we propose a novel XAI framework named PnPXAI, which enables end users to achieve optimal explanations in a plug-and-play manner. PnPXAI addresses the aforementioned challenges of existing XAI frameworks by modularizing it into multiple modules: detector, recommender, explainer, evaluator, and hyperparameter optimizer. As illustrated in Figure \ref{fig:pnpxai_frameweork}, the detector module automatically detects neural network architectures that are used by the recommender to filter the applicable explanation methods. Explanation results from the suggested explainers are then optimized through hyperparameter optimization with the evaluator module. This modular approach ensures that PnPXAI can adapt to a wide range of models, including linear, convolution, recurrent, and transformer modules, as well as complex operations like residual connections. Additionally, PnPXAI supports various input data modalities, including vision, language, time series, and structured data. Consequently, PnPXAI provides users with accurate and reliable explanations, a feature we term \textit{AutoExplanation}.

\subsection{Detector}

The first process undertaken by our framework is the detection of the neural network model architecture. The detector module traces and stores the symbolic graph by iterating over the model provided by end users, which will subsequently be used by the recommender module. The detected symbolic graph not only enables a detailed analysis of the model substructure for selecting applicable attribution methods but also provides layer-wise manipulability, facilitating the automation of layer-specific operations required by relevance propagation methods. If it is not necessary or possible to derive a symbolic graph from the model, such as in the case of tree-based models, the detector module provides basic information about the model architecture.

\subsection{Explainer}

The explainer module, which is a pool of explanation methods to be explored by the recommender module, aims to provide a large set of state-of-the-art methods applicable to a model. Whereas previous work on automatic XAI frameworks \cite{cugny2022autoxai} aimed to fit explanations to specific tasks and user contexts by limiting the scope of applicable methods to model-agnostic ones, such as LIME \cite{ribeiro2016should} and SHAP \cite{lundberg2017unified}, the PnPXAI framework supports both model-specific and model-agnostic methods. These model-specific attribution methods include an equivalent number of methods as non-automatic XAI frameworks \cite{yang2022omnixai}, such as gradient-based methods \cite{srinivas2019full, smilkov2017smoothgrad, adebayo2018sanity}, CAM-based methods \cite{selvaraju2017grad}, relevance propagation methods \cite{bach2015pixel, nam2020relative}, and attention-specific methods \cite{abnar2020quantifying, chefer2021transformer}. This approach has advantages over previous ones in that it provides users with the opportunity to access a wider range of potentially more accurate and reliable explanations. While most preset methods are composed of promising XAI toolkits, such as Captum \cite{kokhlikyan2020captum} and Zennit \cite{anders2021software}, we also provide a flexible generic class for custom methods, allowing them to work seamlessly with the framework.

\subsection{Recommender}

The role of the recommender module is to determine a set of applicable explanation methods by considering input data modalities and model architecture information provided by the detector module. As summarized in Table \ref{tab:mapping_table}, it utilizes a mapping table that consists of two sets: data modalities and neural network architectures. The recommender module supports four types of data modalities—vision, language, structured data, and time series—as well as multi-modalities such as vision and language for VQA tasks. It also supports five types of neural network modules, including linear, convolutional, recurrent, transformer, and decision trees. The module selects candidate explanation methods by intersecting these two sets. For example, if the user-provided model is ResNet50 and the task is VQA, the recommender module suggests nine candidate attribution methods, including LIME, KernelSHAP, Gradient, Gradient $\times$ Input, SmoothGrad, VarGrad, IG, LRP, and RAP. Additionally, the use of the mapping table makes our framework extensible. If users need to implement their custom attribution methods, they only need to add their supported data modalities and model architectures to the mapping table. We believe that this extensibility facilitates advanced usage, such as benchmark studies on new explanation methods.

\subsection{Evaluator}

The evaluator is a module that objectively assesses the plausibility of explanations from various perspectives. As human-grounded evaluations of attribution methods can be misleading and may not accurately measure what a model attends to \cite{adebayo2018sanity}, we evaluate the explanations using quantitative metrics to assess their desirable properties.  
We are inspired by Co-12~\cite{NautaTPNPSSKS23}, which categorizes the evaluation properties; we employ three of them, correctness, continuity, and compactness satisfying the following conditions. 
Criteria requiring human subjective intervention are excluded, and frequently used and distinguishable from other properties are selected.
Similar to the explainer module, our implementation provides a flexible generic class for evaluation metrics, allowing users to incorporate their custom metrics into our framework.

\subsection{Hyperparameter Optimizer}
In addition to the evaluator module, the optimizer module optimizes the selected explanation methods by tuning the set of hyperparameters. This helps mitigate the problem of recommending sub-optimal explanations \cite{yang2022omnixai}, as the selection of hyperparameters significantly affects the quality of explanations \cite{arras2022clevr, cugny2022autoxai}. The module optimizes an explanation on the user-provided dataset through a grid search. It then evaluates the explanation's properties using the selected quantitative metrics. A detailed explanation of the experimental results regarding hyperparameter optimization is provided in the following section, along with Figure \ref{fig:liver_tumor}.

\begin{figure}[!t]
    \centering
\begin{lstlisting}[language=python]
# Load any desired model and dataset to be explained
model, loader = load_model_and_dataloader_for_tutorial('image')

# AutoExplanation
from pnpxai import AutoExplanationForImageClassification

expr = AutoExplanationForImageClassification(
    model=model,
    data=loader,
    input_extractor=lambda batch: batch[0],
    label_extractor=lambda batch: batch[-1],
    target_extractor=lambda outputs: outputs.argmax(-1),
)
\end{lstlisting}
    \captionof{lstlisting}{\textbf{Example code snippet for running AutoExplanation in a plug-and-play manner.} This code demonstrates how to initialize and execute the AutoExplanation, showcasing its ease of integration and use within the PnPXAI framework.}
    \label{lst:autoexp_code}
\end{figure}

\subsection{AutoExplanation}

By processing user-provided inputs with the aforementioned modules, PnPXAI offers \textit{AutoExplanation}, which automatically generates optimal explanations with just a few lines of code, as illustrated in Code \ref{lst:autoexp_code}. By inputting a custom-implemented neural network model and its dataset into the \texttt{AutoExplanation} function provided by the PnPXAI framework and executing it, all user inputs are automatically processed through these modules, and the explanation results are recorded. Illustrative attribution heatmaps for the ImageNet classification task, IMDB movie sentiment analysis task, vision question answering task, credit scoring task, and ECG classification task, achieved by running this \texttt{AutoExplanation} function, are provided in Figure \ref{fig:pnpxai_frameweork}.
\section{Use Cases}


\subsection{Liver Tumor Detection}

\begin{figure*}[!ht]
    \centering
    \includegraphics[width=1.\linewidth]{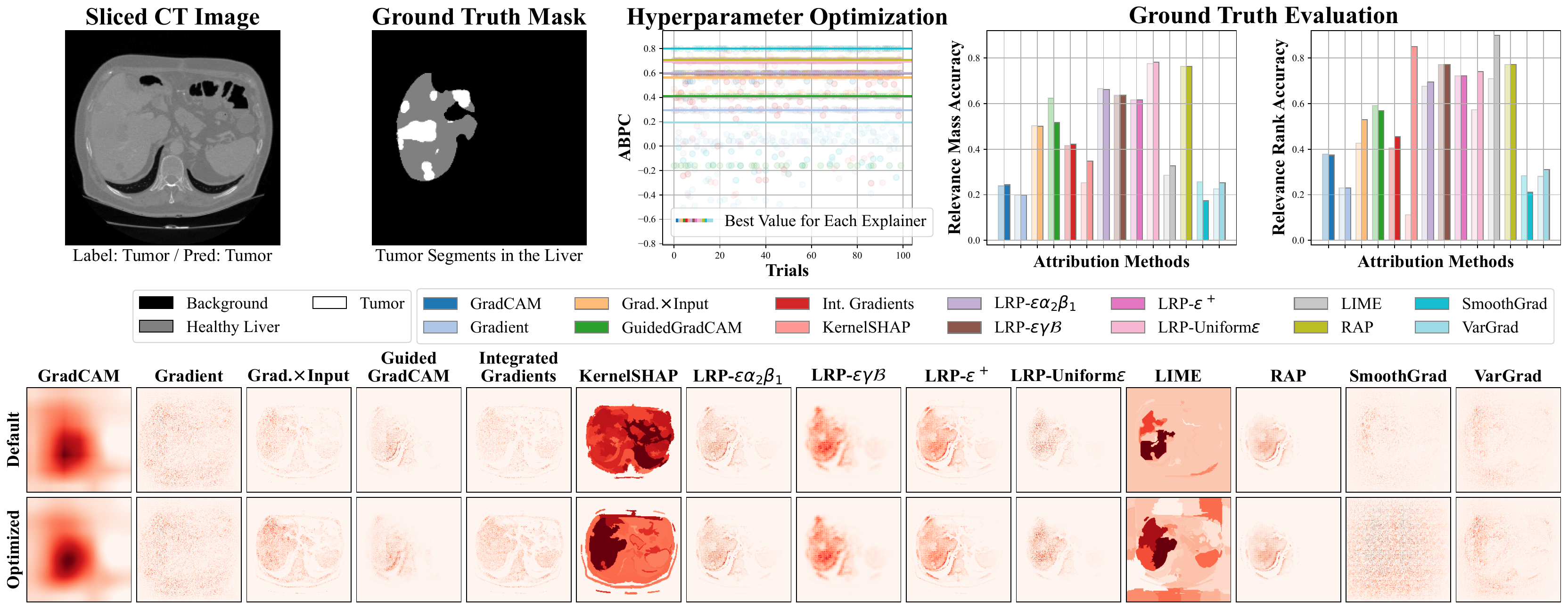}
    \caption{\textbf{Illustration of AutoExplanation for liver tumor detection.} PnPXAI recommends 14 applicable explanation methods and optimizes the selection of hyperparameters for each method on the pre-defined objective, ABPC~\cite{han2023abpc}. This optimization improves relevance accuracy when evaluated against the ground truth segmentation mask. The attribution heatmaps at the bottom rows, where higher attribution scores are indicated by more intense red colors, demonstrate that PnPXAI enables to identify whether the model attributes the segments of the liver.}
    \label{fig:liver_tumor}
\end{figure*}

To demonstrate the practical application of our framework in a medical image classification task, we validate whether the attribution heatmaps highlight the liver segments labeled as tumors in sliced computed tomography (CT) images. To this end, we prepare 2D CT images sliced along the axial axis from 3D CT images, which contain the internal structure of organs. These 3D CT images are obtained from a liver tumor segmentation dataset \cite{bilic2023liver}, which provides ground truth segmentation masks of primary and secondary liver tumors labeled by seven hospitals and research institutions. We train a ResNet50 model on the sliced CT images, where the model outputs ``tumor'' if the input sliced image contains a segment of liver tumor, and ``normal'' otherwise.

By running the \textit{AutoExplanation}, we obtain the optimized attribution heatmaps. As shown in the top center of Figure \ref{fig:liver_tumor}, PnPXAI selects the set of hyperparameters for each explanation method that achieves the best score with respect to the pre-defined objective. ABPC~\cite{han2023abpc} is chosen as the default objective with the evaluation property of correctness as it measures whether the explanation is faithful to the model's decision without requiring the ground truth attribution mask.

As a result of running the \textit{AutoExplanation}, attribution heatmaps from the default and optimized hyperparameters are illustrated at the bottom rows of Figure \ref{fig:liver_tumor}, where the first row and the second row demonstrate attribution heatmaps from the default and optimized hyperparameters, respectively. Notably, most explanation methods attribute the liver segment as identified by the ground truth segmentation masks, despite the presence of various other organs in the input image. Specifically, for relevance propagation-based methods, including variants of LRP and RAP, the attribution heatmaps highlight liver segments and minimally highlight segments of other organs. Additionally, perturbation-based attribution methods, including KernelSHAP and LIME, attribute the liver segment more accurately after hyperparameter optimization.

We also quantitatively analyze whether hyperparameter optimization improves ground truth relevance accuracy \cite{arras2022clevr}, including mass accuracy and rank accuracy. Relevance mass accuracy measures the ratio of the sum of the attribution values within the ground truth mask to the sum of all relevance values over the entire image. Similarly, relevance rank accuracy measures how much of the high-intensity attributions lie within the ground truth. We set both healthy liver segments and liver tumor segments as ground truth attributions, as the model could identify the tumor by recognizing either the segments of the tumor directly or the shape of the healthy liver segments.

The results are depicted on the top right of Figure \ref{fig:liver_tumor}. Hyperparameter optimization increases both relevance mass accuracy and relevance rank accuracy for most of the recommended explainers. Specifically, for perturbation-based attribution methods, including KernelSHAP and LIME, the accuracy improves significantly, implying that the choice of feature mask that creates superpixels is the most important hyperparameter for such methods.

\begin{figure}[!ht]
    \centering
    \includegraphics[width=0.9\linewidth]{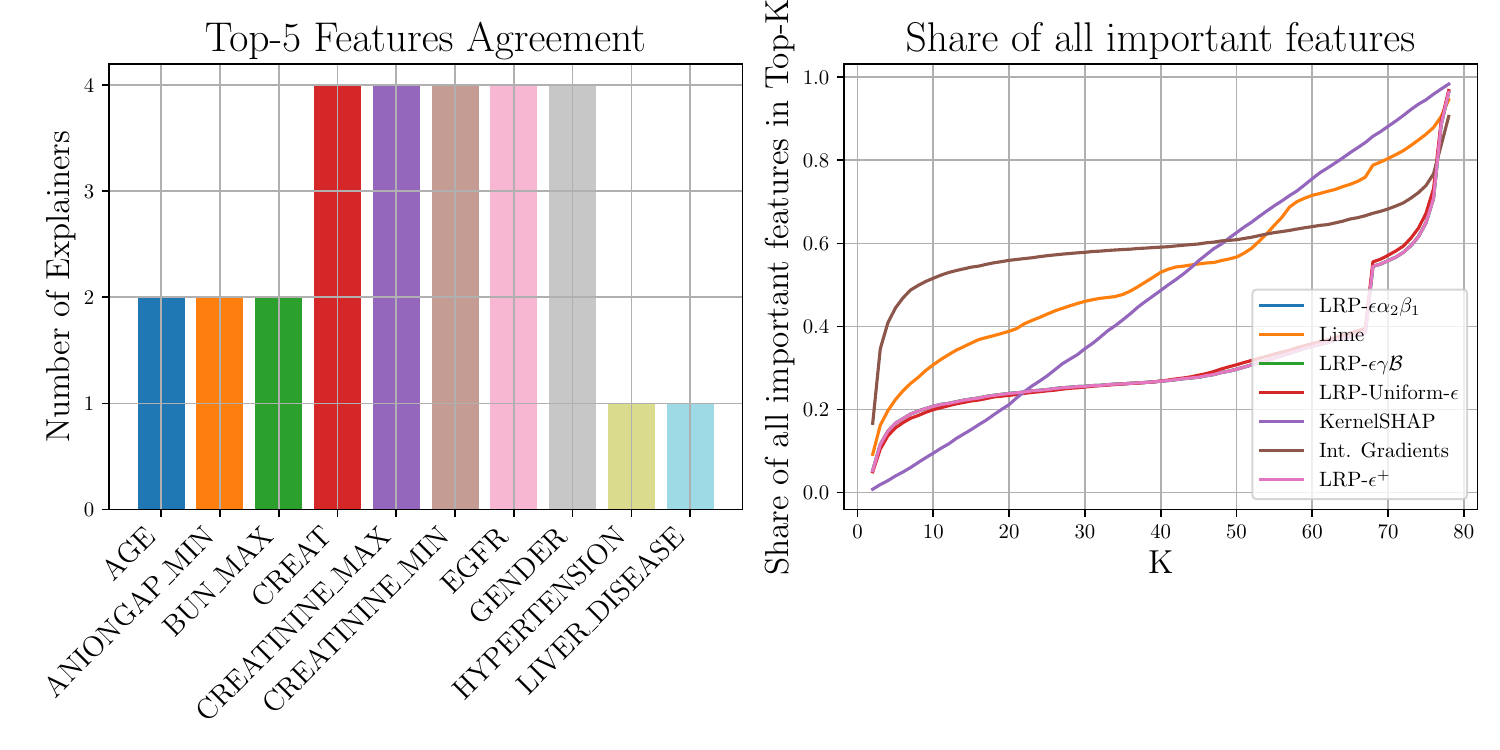}
    \caption{\textbf{Evaluation of PnPXAI in AKI detection.} The top 5 features identified by the selected explanation methods are compared against known AKI biomarkers (eGFR, creatinine, and BUN). The line graph illustrates the alignment of the ABPC metric with the share of expected features among the most attributed ones.}
    \label{fig:aki_detection}
\end{figure}

\subsection{Acute Kidney Injury Detection}
Another use case is acute kidney injury (AKI) detection, where the PnPXAI aims to find insights into the reasons behind the model's decision-making. To test this, we preprocess a MIMIC III \cite{johnson2016mimic} dataset, by extracting 79 features and setting the target to be an existence of AKI in the next 7 days after the patient's admission to the intensive care unit (ICU). Our model is the 8-layer ReLU-activated linear model with 256 hidden neurons in each layer. Following medical studies \cite{makris2016acute}, we identify a list of biomedical markers, proven to be AKI detectors, namely estimated glomerular filtration rate (eGFR), creatinine, and blood urea nitrogen (BUN).

First, we verify the framework's applicability in explaining the model, by analyzing the ability of LRP, Integrated Gradients, LIME, and KernelSHAP to attribute to the most important features. We select the top 5 features and verify that they match the expected markers, having creatinine-related features and eGFR being the most attributed among all explainers, and BUN highlighted by the two explainers (Figure \ref{fig:aki_detection}). Second, we verify the evaluation ability of the framework by identifying the appearance of creatinine-related features, and eGFR in the top-$k$ of the most attributed features, where $k \in \{2..79\}$. The line graph in Figure \ref{fig:aki_detection} depicts better identification of the most important features by LIME, Integrated Gradients, which aligns with the highest scores in ABPC metric. Thus, we verify the model's correctness with the framework's help by detecting the overlap of expected and the most attributed biomedical markers.

\begin{figure*}[!ht]
    \centering
    \includegraphics[width=0.9\linewidth]{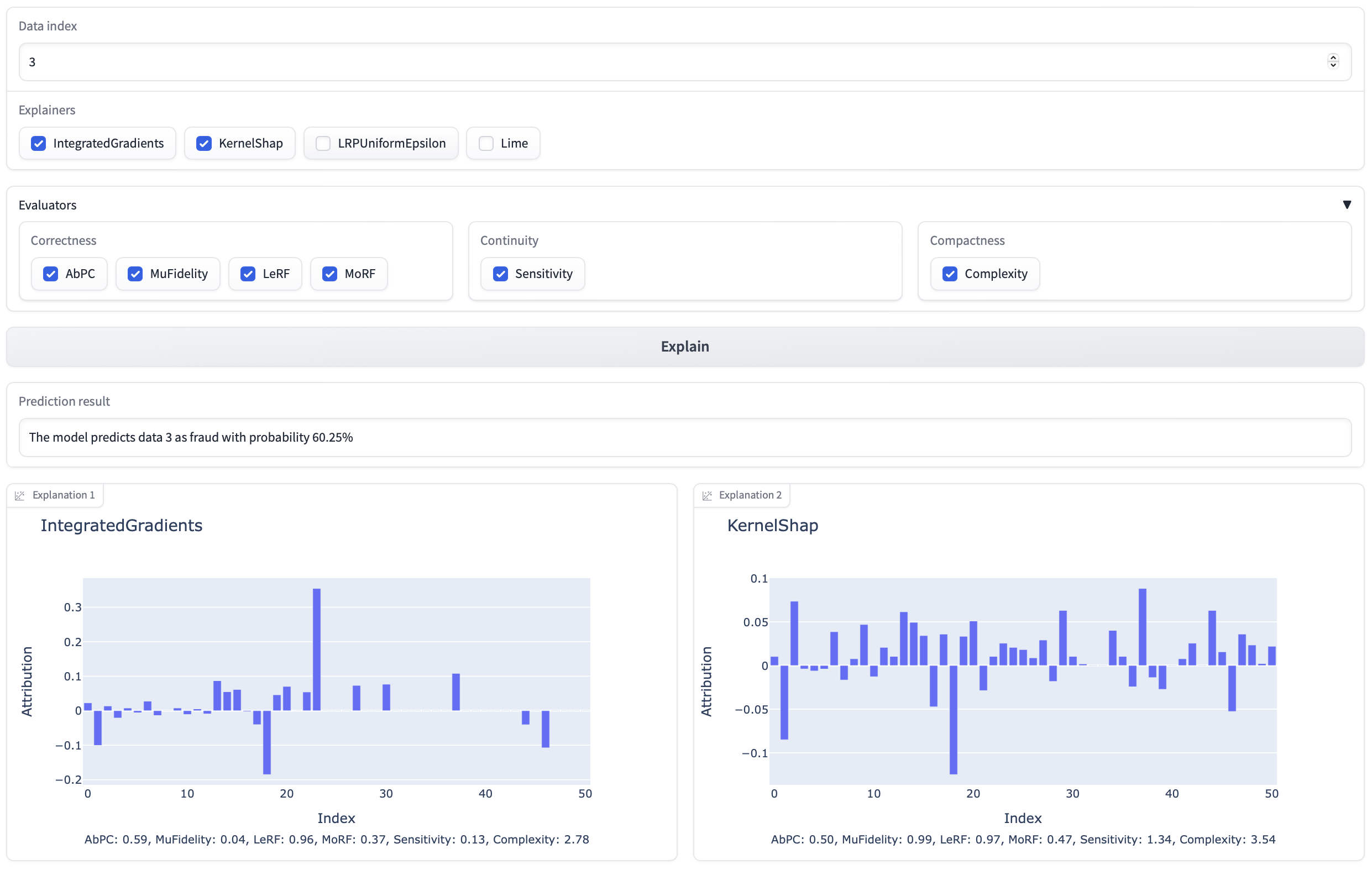}
    \caption{\textbf{Illustration of the interface of a web application for a bank account fraud detection task.} The interface allows users to choose specific explainers and evaluators for given data points. It demonstrates the importance of various features in the model's decision-making process and provides evaluation scores for each explanation algorithm.}
    \label{fig:fraud_detection}
\end{figure*}

\subsection{Bank Account Fraud Detection}

To demonstrate the application of PnPXAI in a real-world scenario, we develop a use case for explaining fraud detection models, a crucial task in the finance sector. This use case focuses on bank account fraud detection, utilizing the Bank Account Fraud (BAF) dataset \cite{jesus2022turning}, which contains information submitted during account opening and corresponding fraud status.
The model is trained to detect whether an account opening request is associated with fraud based on the provided information. We employ various model structures, including ResNet, Logistic Regression, and XGBoost, to showcase the framework's versatility.

Figure~\ref{fig:fraud_detection} illustrates the user interface of the application developed for this use case. The key functionalities, including automatic recommendation of explainers based on the given model structure and calculation of explanation results and evaluation metrics, are easily implemented using the PnPXAI. This ease of use allows developers to create such applications without requiring extensive expertise in XAI, showcasing the framework's potential for practical, user-friendly implementations.

\section{User Survey}

A user survey, designed with the user's perspective at its core, was conducted to validate the effectiveness and convenience of PnPXAI. The user interface for our framework, provided through Gradio~\cite{abid2019gradio}, was tested on ImageNet data for image classification tasks where XAI is the most frequently applied~\cite{NautaTPNPSSKS23}. The survey targeted machine learning/deep learning developers/researchers with direct or indirect experience with XAI, facilitating a reliable evaluation process. A total of 31 participants were recruited from five graduate research groups and one company specializing in AI-driven solutions. 

\begin{table}[!ht]
    \centering
    \begin{tabular}{l||c|c}
    \toprule
        \textbf{Features} & \textbf{Satisfaction $\uparrow$} & \textbf{Importance $\downarrow$} \\ 
    \midrule
        Automatic detection & 4.06 $\pm$0.15 & 2.35 $\pm$0.21 \\ 
        Recommendation & \textbf{4.13 $\pm$0.13} & \textbf{1.90 $\pm$0.16} \\ 
        Hyperparameter opt. & 3.94 $\pm$0.16 & 2.32 $\pm$0.22 \\ 
        Evaluation & 3.97 $\pm$0.16 & 2.26 $\pm$0.20 \\ 
    \bottomrule
    \end{tabular}
    \caption{\textbf{Average user satisfaction score and importance rank for each PnPXAI feature.} Satisfaction is assessed using a 5-point Likert scale, where users rated the convenience and usefulness of each feature. Importance is evaluated by ranking the four key features of PnPXAI in order of their significance in distinguishing PnPXAI from other XAI tools.}
    \label{tab:user_study}
\end{table}

The survey consists of questions regarding user experience with XAI algorithms and questions assessing satisfaction with each PnPXAI feature. As presented in Table~\ref{tab:user_study}, the feature that automatically recommends applicable XAI algorithms received the highest satisfaction score, though others also demonstrated high satisfaction, with scores close to 4.

Among the 31 participants, 27 (87.1\%) had direct experience using XAI algorithms. When asked about the challenges of using these tools, 70\% of them responded that ``it is difficult to trust explanations", followed by ``it is difficult to understand explanations" (51.9\%) and ``it is difficult to find and apply XAI algorithms" (48.1\%).

According to user feedback, PnPXAI enhances reliability by aligning with multiple metrics for each of the three categorized evaluation properties and provides more accurate explanations through hyperparameter optimization of XAI methods. Moreover, it increases time efficiency by automatically detecting model structures and recommending \& applying applicable XAI methods.

\section{Conclusion}

In this paper, we introduced PnPXAI, a universal framework designed to address the limitations of current XAI frameworks, such as their inflexibility to diverse model architectures and data modalities, lack of easy-to-use evaluation and optimization, and barrier to utilizing various explanation algorithms at its most. By modularizing the framework into detector, recommender, explainer, and evaluator modules, PnPXAI offers a comprehensive solution that supports diverse data modalities and neural network architectures in a plug-and-play manner. We validated the usefulness of each key functionality of PnPXAI through a user survey of 31 participants. Additionally, we demonstrated its ability to provide accurate and reliable explanations through ground truth evaluation in practical use cases across various domains, including medicine and finance.

For future work, we aim to expand PnPXAI’s capabilities to include explainers for generative large language models (LLMs). With its proven, easy-to-integrate design, we anticipate collaboration with the global AI research community to develop state-of-the-art explanation methods, enhancing the interpretability and trustworthiness of LLMs.

\bibliography{references}

\end{document}